\documentclass[10pt,twocolumn,letterpaper]{article}

\usepackage{cvpr}              

\usepackage{graphicx}
\usepackage{amsmath}
\usepackage{amssymb}
\usepackage{booktabs}
\usepackage{textcomp}
\usepackage{xcolor}
\usepackage{subcaption}
\usepackage{tikz}

\usepackage{color, colortbl}
\definecolor{LightCyan}{rgb}{0.85,1,1}
\definecolor{Gray}{gray}{0.95}

\def\BibTeX{{\rm B\kern-.05em{\sc i\kern-.025em b}\kern-.08em
    T\kern-.1667em\lower.7ex\hbox{E}\kern-.125emX}}

\usepackage[pagebackref,breaklinks,colorlinks]{hyperref}

\usepackage[capitalize]{cleveref}
\crefname{section}{Sec.}{Secs.}
\Crefname{section}{Section}{Sections}
\Crefname{table}{Table}{Tables}
\crefname{table}{Tab.}{Tabs.}

\def\cvprPaperID{*****} 
\def\confName{CVPR}
\def\confYear{2023}

\begin{document}

\title{Entropy Transformer Networks: A Learning Approach via \\Tangent Bundle Data Manifold}

\author{Pourya Shamsolmoali$^1$, Masoumeh Zareapoor$^2$\\
$^1$East China Normal University  \; \; \; \;
$^2$Shanghai Jiao Tong University
}
\maketitle

\begin{abstract}
This paper focuses on an accurate and fast interpolation approach for image transformation employed in the design of CNN architectures. Standard Spatial Transformer Networks (STNs) use bilinear or linear interpolation as their interpolation, with unrealistic assumptions about the underlying data distributions, which leads to poor performance under scale variations. Moreover, STNs do not preserve the norm of gradients in propagation due to their dependency on sparse neighboring pixels. To address this problem, a novel Entropy STN (ESTN) is proposed that interpolates on the data manifold distributions. In particular, random samples are generated for each pixel in association with the tangent space of the data manifold and construct a linear approximation of their intensity values with an entropy regularizer to compute the transformer parameters. A simple yet effective technique is also proposed to normalize the non-zero values of the convolution operation, to fine-tune the layers for gradients' norm-regularization during training. Experiments on challenging benchmarks show that the proposed ESTN can improve predictive accuracy over a range of computer vision tasks, including image reconstruction, and classification, while reducing the computational cost.
\end{abstract}

\begin{figure}[t]
\includegraphics[width=0.95\columnwidth]{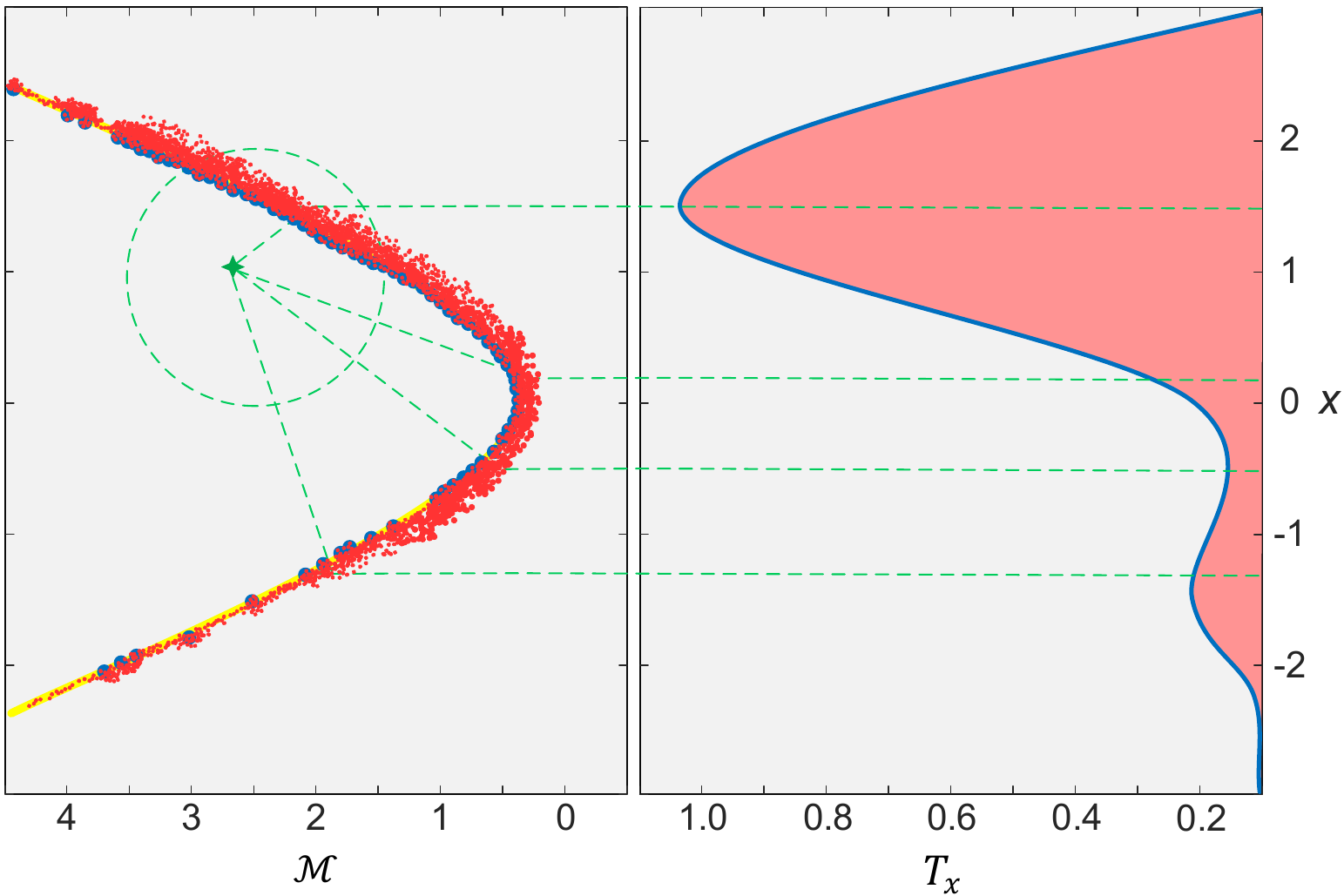} 
\centering
\caption{Data transformation and resulting distributions. Left: Data manifold $\cal{M}$ (yellow curve), context points (blue dots), interpolate points (red dots), Green star is the model observation. Stationary parts in target density (green lines). Right: Plot of the transformed density (blue curve), and histogram of the projected samples (red).}   
\label{fig:1}
\end{figure}

\section{Introduction}
\label{sec:intro}

Determining discriminant features is challenging in various computer vision tasks, such as object detection \cite{zhang2018single} and image retrieval \cite{noh2017large}. To detect data relations we need an image descriptors to efficiently encode image features that are invariant to geometric image transformations. For example, to detect objects in real-world images, we have learned that the appearance of an object in an image depends on different types of deformations, such as rotations and scale changes. However, the features of an object should remain consistent (invariant) to these transformations. Generally, CNNs do not make use of prior learning to deal with the transformation invariance of objects across different domains. As a substitute, CNNs use data augmentation to integrate such priors during training. To some extent, data augmentation improves CNN performance, but provides no assurance that transformation invariance will be achieved for the data that has never been seen throughout the training. In \cite{gerken2022equivariance}, the authors investigate the role of rotational equivariance in CNNs.

In addition to data augmentation, group equivariant CNNs \cite{cohen2016group}, \cite {cohen2018spherical} have been proposed to constantly address particular transformations of the input. For example, in \cite{he2021efficient}, a CNN is integrated with equivariant self-attention layers, to separate the feature aggregation process into a kernel generator, and decouple the spatial and additional geometric dimensions in the computing. 
Moreover, Jaderberg et al. \cite{jaderberg2015spatial} have proposed the STN, a discernible module that uses an attention mechanism to perform spatial manipulation and transformation on the image by a CNN. 
In contrast to pooling layers, in which the receptive fields are constant, the STN is an active process that uses the attention mechanism to transform the feature map of the input image by producing a transformation for the image. In \cite{worrall2017harmonic}, a rotation and translation approach is proposed, which generalizes poorly to the composition of feature transformations, because this model does not process the local information in complex scenarios.

Indeed, the above models \cite{jaderberg2015spatial}, \cite {worrall2017harmonic} are based on bilinear interpolation for calculating the values of pixels. Bilinear interpolation is local, which is one of its weaknesses, as it only leverages the four nearest pixels of the query. Since bilinear interpolation does not consider high-level transformation, the performance of deep neural networks are reduced when the image size changes. Among the techniques used for predicting missing pixels, linear interpolation has attracted widespread attention due to its simplicity. However, this technique cannot perform accurately because of redundant noise. The limitation of the bilinear interpolation was addressed in \cite{jiang2019linearized} by proposing Linearized-STNs (L-STNs). This approach has better performance compared to \cite{jaderberg2015spatial}, however, in L-STNs random noise is generated during training, over-fitting often occurs, and linear regression is inefficient.

STNs are generally adopted in CNNs as pre-alignment modules for different tasks \cite {lin2017inverse}. Current STN approaches cannot deal with extensive downsampling to increase the network capacity. To address this problem, we propose a Entropy STN (ESTN), where the interpolation method allows to increase the capacity of STN for learning invariance against different type of transformations. First, a local tangent space approximation \cite{yang2005better} is performed to estimate the data tangent space by using the variational autoencoders (VAEs) \cite{kingma2013auto}, and enforce the interpolation module to generate auxiliary samples in the neighborhood of the query pixel. Adopting an entropy regularization allows exploring more samples in the search space along the data manifold and improving the convergence properties. This allows learning the geometric correlations that lie at the intersection of the input image and the geometric manifolds. The space of geometric manipulations is based on the transformed image (see Fig. \ref{fig:1}), and samples are formed in the transformed domain, where their spatial locations are determined by the transformation. Therefore, gradient features are transformation-aware to improve network performance by identifying the important regions of an image and detecting the changes in accordance with the transformation.

To reconstruct an image, we rely on the content information in the local tangent space, followed by smoothing of the orthogonal direction factors in the local tangent space. Smoothing the orthogonal direction factors allows to maintain the local geometries of the surface that are linearly approximated in the tangent space. The ESTN improves the learning ability of deep neural networks, which in turn can be used to promote any equivariance translation in CNN architectures. To incorporate the proposed ESTN into any CNN architecture and have smooth training, we introduce a norm-regularization method by retaining the same variance of the weighted gradients in the layers. Our method does not require singular value decomposition to process the non-zero values of a CNN on the underlying data manifold. With the above modifications, our model achieves higher accuracy with lower computation costs on different tasks.

\section{Related Work}

\noindent{\bf Equivariant CNNs.}
Recently, there has been a strong interest from the machine learning community in the design of CNN models that are equivariant in terms of image transformation. Equivariance in terms of transformation is possible via transforming CNN filters or manipulating feature maps \cite{wang2022approximately, cohen2016group}. In \cite{zhu2022scaling}, an equivariant CNN to space and scaling is proposed, which is equivariance for the regular representation of scaling and translation. Moreover, \cite{sangalli2022moving} proposed an efficient equivariant network by reducing the computation of moving frames at the input stage rather than repeating computations at all layers. 
In fact, these approaches support limited transformations because the computational cost increases linearly with the cardinality ratio of the transformation.
The equivariance approaches with regard to uninterrupted transformation are divided into two classes: (1) those that increase the input in a guiding foundation \cite{cohen2018spherical}, and (2) those that assess convolutions through a comprehensive geodetic system \cite{henriques2017warped}. \cite{zhong2021t} introduces an equivariant hypergraph module that can be adopted in various neural networks. \cite{satorras2021n} also recently proposes efficient equivariant graph neural networks that can easily scale to higher-dimensional spaces. In addition, a permutation-equivariant network is proposed in \cite{kim2022equivariant} to capture both global and channel-wise contextual information. 
In \cite{singh2022learning}, to build an equivariant network, orthogonal moments of the function are used as an effective means for encoding global invariance with respect to rotation and translation in fully-connected layers of the network. 

A new class of equivariant CNNs was proposed in \cite{cohen2019gauge}, based on data manifold learning and intrinsic geometry. In \cite{sutanto2020learning}, to address the problems of learning representations, the authors proposed a manifold neural network model that learns from a level-set function of the constraint. However, these methods can not realize the relative feature-space relationship between two transformed encoded inputs. For example, the relative feature space relationship between two transformed images. \\
\vspace{-8pt}

\noindent {\bf Spatial Transformer Networks (STNs)}. is a method to integrate the learned image deformation into a CNN architecture. A ST layer consists of a sub-network to predict parameters with a deformation function. In \cite{lin2017inverse}, Inverse Compositional STNs (IC-STNs) was proposed as an alignment framework by adapting the standard Lucas-Kanade (LK) algorithm into STNs.
In this model, the image coordinate $x\in R^2$ and the input intensity at the same coordinate $U_{(x)}\in R^C$ ($C$ denotes the number of channels) are used to estimate the transformation parameter $\theta$ as $\mathcal{T}_{\theta}(x)$ and the transformed output $V_{(\tilde x)}$ at the target coordinate $\tilde x$. 
\begin{equation}
V_{(\tilde x)}=\sum_x U(\mathcal{T}_{\theta} (x)) K(\tilde x, \mathcal{T}_{\theta} (x)),
\label{eq:1}
\end{equation}
in which $K(., .)$ is the convolution's kernel to reflect the impact of this process on each pixel. In this model, the kernel influences the whole image and creates gradient values for all the pixels. This process involves back-propagation throughout the pixels in the input image and to all the pixels in the transformed one, which is expensive. The $K$ is adjusted in bilinear interpolation, thus, $K(x, y)=0$ where $x$ and $y$ are not neighbours. For this reason, the gradient just flows over sub-pixels gradient which reduce the model performance under down-sampling. Moreover, the gradient of the sub-pixels cannot address the sweeping changes that raised upon the changes of transformation parameter $\theta$, because it is impossible to capture minor transformations by direct neighbouring pixels. Therefore, the intensity values with non-zero gradients from $K$ need to be selected in such a manner that is invariant to deformations, in order to accommodate how the image would be change under deformations. Our ESTN has been designed to address different variations of geometry transformations by interpolating on the data manifold distribution.

\section{Entropy Spatial Transformer Network}
In this section, we illustrate our proposed ESTN. We first discuss the manifold learning, before introducing our model.
\subsection{Interpolation on Data Manifold}
In our model, we use the VAE, which is a strong generative model that contains two modules: (1) an encoder, which is a recognition network to learn the latent manifold representation of the data in the input space; (2) the decoder, which learns how to reconstruct the data from the latent space. A prior distribution is defined for the latent manifold representations, and a mapping function is used to generate a surface in input space. In general, each $x$ has an underlying manifold with a lower dimension represented as ${\cal{M}}\cong R^C$. We approximate Jacobian propagation to extend the input data perturbed in the Jacobian order. In terms of regression, this strategy can be implemented by generating auxiliary samples $z$ over the underlying manifold $\cal{M}$. Therefore, due to the smoothness of the data on the manifold and the smooth inductive bias of neural networks, realistic results can be obtained by perturbing each input pixel with a Gaussian random vector and marginalizing the samples $z$ in the context of linear regression. In this work, for each pixel, we generate auxiliary noise $R_z$ within $\cal{M}$ and the key point is how we ingeniously estimate $\cal{M}$ and design $R_z$ to compel the intended smoothness. Therefore, linear regression is used to find $w$ for minimizing $\Vert z-xw\Vert^2 \in R^{(\hat c)}$ ($xw$ is an inner product so $C = \hat{c}$ and is a 2D matrix). To retain the input dimension, we use Bernoulli probability $(p)$, therefore, the input $x$ represented as $(r\circ x)$ in which $r\in \{0,1\}$ is a random matrix. To marginalize $\mu$ the samples, we formulate the following function
\begin{equation}
\min_r  \mu_{r\sim (p)}[\Vert z-(r\circ x)\Vert^2].
\label{eq:2}
\end{equation}
where it returns a set of possible alignments compatible with $x$ drawn from the manifold distribution and this allows to efficiently obtain both local and contextual structures. To form $R_{\cal M}$ on the basis of the Jacobian, we use
\begin{equation}
R_{\cal{M}}=\mu_{x \sim dis(x)} \sum_{v\in T_x \cal{M}} \Vert J_x . v \Vert,
\label{eq:3}
\end{equation}
where, $dis(x)$ is the true data distribution, $T_x \cal{M}$ denotes the tangent space of $\cal{M}$ at $x$, $J$ is the Jacobin. These parameters are used to regularize the manifold's smoothness in view of the norm of its Jacobian and the perception of the tangent space. This specifies a space $m_x (s_1),...,m_x (s_c )\in T_x \cal{M}$, in which $s_i$ denotes a standard space of $R^C$. The linear approximation around $x$ and close to $T_x\cal{M}$ creates a broader context to the local transformation, and enhances the robustness of the network. Therefore, to compute the transformation parameter $\mathcal{T}_{\theta}$ of a point $x_i $, the local linear interpolation $L_{(x)}$ on the tangent subspace is calculated as
\begin{equation}
L_{(x)}=\{U(\mathcal{T}_{\theta}({x_i})) + l_i(\mathcal{T}_{\theta}(x) - \mathcal{T}_{\theta}(x_i)) \}.
\label{eq:4}
\end{equation}
where $l_i$ is a matrix that stands for the local linearization over the underlying data manifold that we look for. To accurately perform interpolation, we should define a local chart around $x$ which is compatible with the nature of the data manifold.
Indeed, ${\mathcal T}_{\theta}$ needs to be very sensitive to changes, for example, from point $x_i$  to another transformed point $x_j$, but insensitive to any other changes. To obtain $l_i$, the local linearization should be performed at the transformed coordinate ${\mathcal T}_{\theta}({\cal M}_{x_i})$, while ${\mathcal T}_{\theta} ({\cal M}_x)$ is invariant. Thus, $l_i$ in (\ref{eq:4}) represents the gradient of  $L_{(x)}$ in terms of $x$ on the tangent manifold and can be expressed by an orthogonal basis. To acquire $l_i$, firstly, we perform interpolation near the point $x_i$ on the underlying data manifold $T_{x_i}\cal M$, $\{x_i^k \sim N(x_i, \sigma)\}$ where $\{k\in (1,2,…,K) \}$, in which $N(x_i, \sigma)$ is the data distribution derived from $x_i$ with standard deviation $\sigma$. With the weighted least-squares method \cite{li2018image}, we can produce sample points in the $N \times N$ neighborhood of $x_i$ and estimate the nearest points. In the weighted least-squares, we set a weight to each point $x$ in the $N \times N$ neighborhood of $x_i$ in accordance with the distance between $x$ and $x_i$. The weight on $T_{x_i} \cal M$ is computed by
\begin{figure}[t]
\centering
\includegraphics[width=0.96\columnwidth]{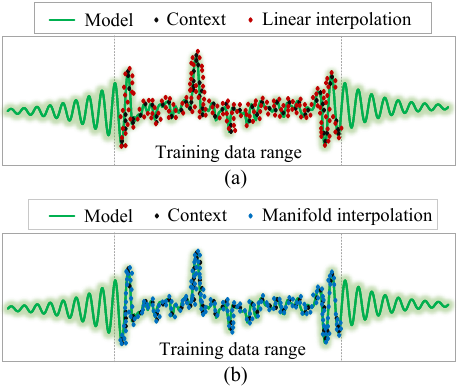} 
\caption{Interpolation methods. (a) Sample generation by linear interpolation. (b) Sample generation based on data manifold.}   
\label{fig:3}
\end{figure}

\begin{equation}
W_i x= \frac{1+\frac{N-1}{2}\sqrt{\frac{N-1}{2}} - \sqrt{(x-x_i)^2}}{\frac{N-1}{2}\sqrt{\frac{N-1}{2}}}
\label{eq:5}
\end{equation}
We streamline the notation $L({\mathcal T}_{\theta}({\cal M}_{x_i^k}))$ as $L_i^k$, and ${\mathcal T}_{\theta} ({\cal M}_{x_i^k })$ as $\tilde x_i^k$, where $\tilde x_i^k=[\tilde u_i^k, \tilde v_i^k]$, is based on the LK \cite{chang2017clkn}, therefore we create two data matrices for the sampled points as follows
\begin{equation}
y_i= x_i l_i \vert l_i \in R^C ,
\label{eq:6}
\end{equation}
\begin{equation}
y_i=
\begin{bmatrix}
L_i^1 - L_i^0 & L_i^2 - L_i^0 & ...& L_i^{k-1}- L_i^0
\end{bmatrix}^{ER}
\label{eq:7}
\end{equation}
\begin{equation}
x_i=
\begin{bmatrix}
\tilde u_i^1 - \tilde u_i^0 & \tilde u_i^2-\tilde u_i^0 &  & \tilde u_i^{k-1}-\tilde u_i^0 \\
\tilde v_i^1-\tilde v_i^0 & \tilde v_i^2-\tilde v_i^0 & ... & \tilde v_i^{k-1}-\tilde v_i^0  \\
1 & 1 & & 1
\end{bmatrix}^{ER}
\label{eq:8}
\end{equation}
To estimate the displacement in the pixels, we analyse the changes in pixels by using the intensity gradients of the image in all the pixels. Therefore, $l_i$ can be computed by weighted least-squares $W_i$ on data manifold $T_{x_i} \cal M$ as
\begin{equation}
l_i=(x_i^{ER} W_i x_i + \xi A)^{-1} x_i^{ER} W_i y_i
\label{eq:9}
\end{equation}
where $A$ denotes a $s \times s$ matching matrix, and $\xi$ is a distribution with mean vector 0. To solve the nonconvex problem of Eq. (\ref{eq:9}), generally the Tikhonov method is use as a regularizer \cite{jiang2019linearized}. However, the performance of this method is sensitive to the choice of parameter $\gamma$. In our model, projected gradient descent is used, whereas both the gradient and the projections relied on the basis of the Tikhonov metric. By taking $\frac{1}{\gamma}$ as the learning rate, the projected gradient is equivalent to solve the non-convex problem with the entropy regularizer (ER) \cite {benamou2015iterative}: 
\begin{equation}
V(\tilde x)=\sum_u U(\gamma {\mathcal T}_{\theta} (x)) K(\tilde x, {\mathcal T}_{\theta} (x))+\gamma H({\mathcal T}_{\theta})
\label{eq:10}
\end{equation}

in which $H({\mathcal T}_{\theta})= \sum_{i, j} {\mathcal T}_{i j} \log {\mathcal T}_{i j}$, here ${\mathcal T}_{i j}$ is an element of ${\mathcal T}_{\theta}$ and represents the probability that sample point $i$ matches $j$. By replacing the Tikhonov estimation with the ER, we derive the weight of the regularizer which is a significant hyperparameter to improve the convergence. Fig. \ref{fig:3}, shows the performance of our proposed interpolation model in comparison with linear interpolation.  In our model, the generated samples are in line with the data samples (context) and the data manifold.
\subsection{Network Regularization}
In this section, we define the proposed transformation layer and the essential properties. To execute a transformation, the sampler needs to take a group of sampling points along with $U_{(x)}$ to generate the output feature map $V_{(x)}$. To perform such transformation, we project the proposed ESTN layer kernel into the group of norm preserving kernels by defining its differentiable values. We will discuss how to set the convolution layer to maintain the norm via singular value regularization without dependency on the decomposition of the values. 

More precisely, in a convolution layer, let $k$, $c$, and $d$ be the kernel size, input channels, and output channels respectively, therefore the gradient is formulated as
\begin{equation}
\nabla_u=\hat W \nabla_v,
\label{eq:11}
\end{equation}
in which $u\in R^c$  is a c-dimensional vector, $\nabla_u$ denotes the input's gradient, $v\in R^d$ is the $d$-dimensional vector, and $\nabla_v$ denotes the output's gradient of the convolution. $\hat W$ is an $c\times d$-dimensional matrix which performs backpropagation for the convolution layer. This transformation illustrated as
\begin{equation}
\begin{array}{lr}
\nabla_u=\sum_{i=1}^c \psi_i p_i <\nabla_v, q_i>, \\
\nabla_v=\sum_{i=1}^c q_i <\nabla_v, q_i >,
 \end{array}
        \label{eq:12}
\end{equation}
in which $\psi_i$ is a singular value of $\hat W$  and $p_i$, $q_i$ are corresponding left and right singular vectors, respectively. Therefore, the estimated values of the norms of gradients are calculated as
\begin{equation}
\begin{array}{lr}
       E[\Vert \nabla_u\Vert_2^2] =\sum_{i=1}^c \psi_i^2 E[c ({\cal M})\vert < \nabla_v, q_i > \vert^2 ],\\
       E[\Vert \nabla_v\Vert_2^2] =\sum_{i=1}^d E[d ({\cal M})\vert < \nabla_v, q_i > \vert^2 ],\\
        \end{array}
        \label{eq:13}
\end{equation}
in which, for the equal data range, we have: $p_i^T p_j=q_i^T q_j=1$ if $i=j$, otherwise 0. Therefore, to preserve the gradients' norm, we propose to have $E [\nabla_u] = E [\nabla_v]$, by assigning non-zero singular values to $\psi$. It can be obtained by
\vspace{-3mm}
\begin{equation}
\psi^2= \frac{\sum_{i=1}^d E[\vert <\nabla_v, q_i >\vert^2]}{\sum_{i, \psi_i} E[\vert <\nabla_v, q_i >\vert^2]}
\label{eq:14}
\end{equation}
\begin{figure*}[t]
\centering
\includegraphics[width=1.89\columnwidth]{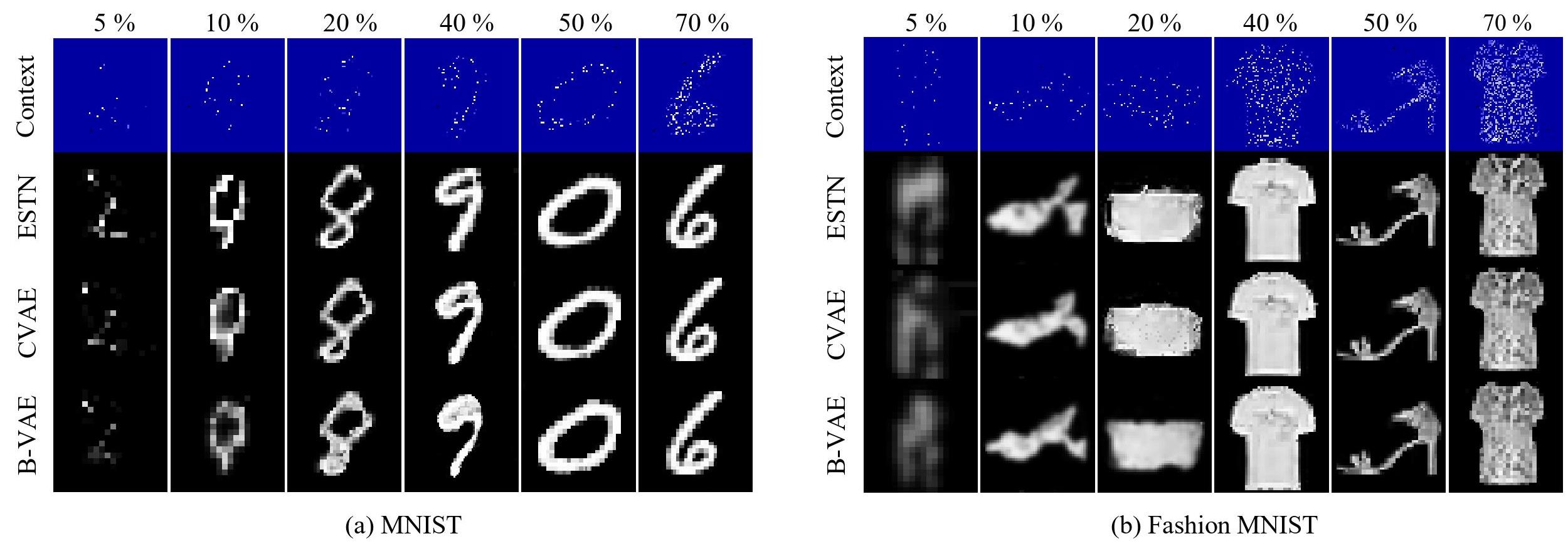} 
\caption{Marginal Log-likelihood and qualitative results on MNIST, and Fashion MNIST. The images stand for the context points (pixels quantity) (first row), ESTN target predictions (second row), CVAE target predictions (third row), and $\beta$-VAE target predictions. Each column represents a specific percentage of the context distribution.}   
\label{fig:5}
\end{figure*}
in which the denominator aggregation in a singular vector $q_i$ are based on the non-zero singular values, $\psi_i \neq 0$. The ratio in Eq. (\ref{eq:14}) is the ratio of the estimated gradient of $\nabla_v$ divided by the total gradient that never fall in the null-space, or kernel of the matrix $\hat W$, so we can approximate this assumption as $\frac{d}{\min(d, c)}$ . Based on this assumption, $\frac{\min(d, c)}{d}$ of the gradient of $\nabla_v$ will fall in the $\min(d, c)$-dimensional subspace, for basis $\{q_i\vert \psi_i\neq 0\}$. Therefore, we can regularize the norm by changing the singular values ($\neq 0$) to $\sqrt{\frac{d}{\min(d, c)}}$. However, implementing this values $\frac{d}{min(d, c)}$ is computationally expensive, due to the square root of matrix and  if not being well implemented, it can delay the training process. 
To mitigate this issue, we use an iterative algorithm to compute the square root of matrix with the help of matrix multiplication. Inspired by \cite{higham1997stable}, this iterative process is presented as: 
\begin{equation}
\begin{array}{lr}
Y_{k+1} = \frac{1}{2} Y_k(3I-Z_k Y_k),\\
Z_{k+1}=\frac{1}{2}(3I-Z_k Y_k)Z_k,
 \end{array}
        \label{eq:15}
\end{equation}
where $k$ is the iteration number. It has been shown that, $Y_k$ and $Z_k$ converge to the $A^{\frac{1}{2}}$ and $A^{-\frac{1}{2}}$ respectively. Therefore, as the iterations only entail matrix multiplication, we can efficiently implement them on GPUs.

\begin{table*}
\centering
\caption{Marginal likelihood evaluation on five datasets.}
\small{
\begin{tabular}{@{}|l |c|c|c|c|c|c|@{}}     \hline
\rule{0pt}{1\normalbaselineskip} Model & Param & MNIST & F-MNIST & SVHN & CelebA & ZSMM \\    \hline\hline
\rule{0pt}{1\normalbaselineskip} VAE & 210K  & 1.08 $\pm$ 0.04 & 2.29 $\pm$ 0.02 & 3.76 $\pm$ 0.03 & 2.92 $\pm$ 0.04 & 0.48 $\pm$ 0.02  \\   
$\beta$-VAE \cite{chen2018isolating} & 415K  & 1.14 $\pm$ 0.03 & 2.36 $\pm$ 0.01 & 3.88 $\pm$ 0.02 & 3.08 $\pm$ 0.02 & 0.61 $\pm$ 0.01  \\               
CVAE \cite{ramchandran2022learning} & 433K & 1.23 $\pm$ 0.02 & 2.41 $\pm$ 0.02 & 3.91 $\pm$ 0.04 & 3.13 $\pm$ 0.02 & 0.74 $\pm$ 0.03  \\   \rowcolor{LightCyan}  
ESTN (ours) & 219K  & 1.28 $\pm$ 0.01& 2.47 $\pm$ 0.00 & 3.95 $\pm$ 0.01 & 3.22 $\pm$ 0.01 & 1.12 $\pm$ 0.01   \\   \hline
\end{tabular}
 \label{tab:1}
}
 \end{table*}

\begin{figure}
\centering
\includegraphics[width=0.97\columnwidth]{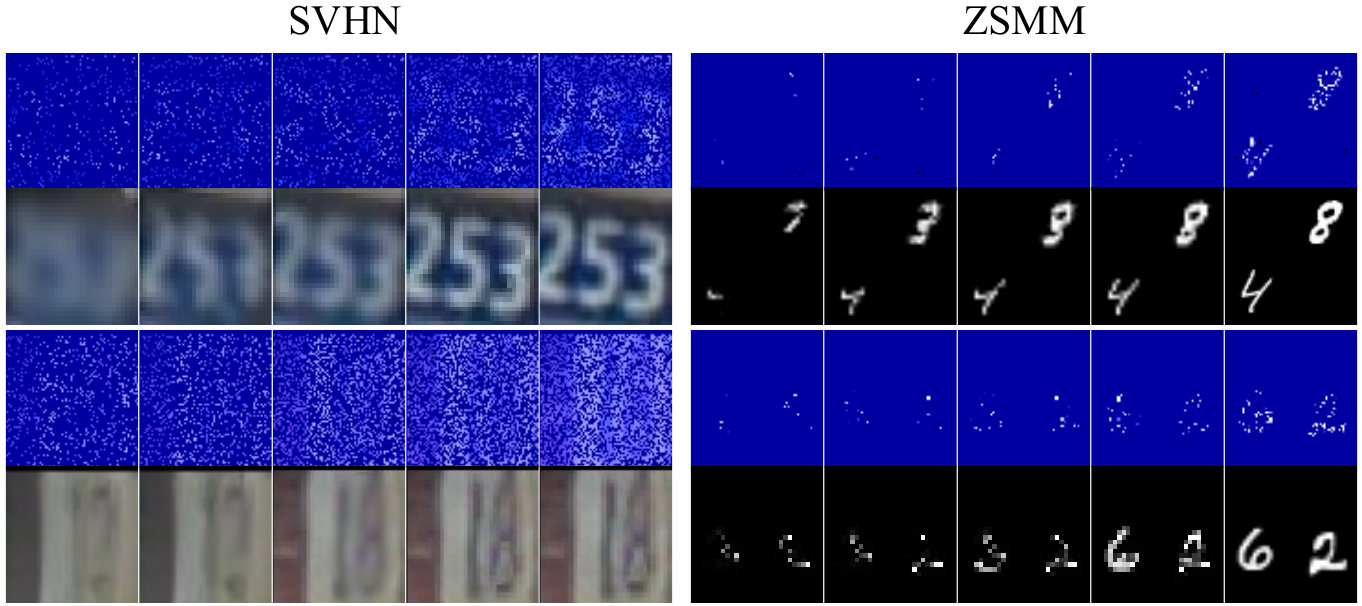} 
\caption{Performance evaluation. From each dataset a random image is selected and different quantity of noise is generated. The first rows from left to right contain the noisy images (5\%, 10\%, 20\%, 30\%, 50\% of random pixels). The second rows show the performance of ESTN in image completion.}   
\label{fig:4}
\end{figure} 
\begin{table*}
\small{
\centering
\caption{Classification errors (\%) of different models with various numbers of data on Fashion MNIST, SVHN, and CelebA.}
\begin{subtable}[t]{0.47\textwidth}
\begin{tabular}{|l |c |c|c|}     \hline
\rule{0pt}{1\normalbaselineskip} Method & 1k labels & 3k labels & 5k labels  \\    \hline\hline
\rule{0pt}{1\normalbaselineskip} CNN & 14.47 $\pm$ 0.23 & 12.63 $\pm$ 0.24 & 9.52 $\pm$ 0.15 \\         
 VAE  & 14.21 $\pm$  0.30 & 12.49$\pm$  0.18 & 9.46 $\pm$  0.11  \\   
 STN \cite{jaderberg2015spatial} & 11.84$\pm$  0.21 & 9.37$\pm$ 0.05  & 6.76 $\pm$ 0.06   \\  
 GCNN \cite{singh2022learning} & 9.63 $\pm$ 0.10 & 7.08 $\pm$ 0.19  & 4.49 $\pm$ 0.15   \\   
 L-STN \cite{jiang2019linearized}  & 8.92 $\pm$  0.12 & 6.83 $\pm$  1.24  & 4.24 $\pm$  1.08   \\   
 SE3Mov \cite{sangalli2022moving} & 8.07 $\pm$  0.24 & 4.87 $\pm$  0.27  & 2.83 $\pm$  0.15   \\  
 ScDCFNet \cite{zhu2022scaling} & 7.98 $\pm$ 0.09 & 4.72 $\pm$  0.12  & 2.65 $\pm$  0.09   \\ \rowcolor{LightCyan}
 ESTN (ours) & {\bf 7.21$\pm$  0.06} & {\bf 4.43 $\pm$  0.8}  & {\bf 2.19 $\pm$  0.16}   \\  \hline
\end{tabular}
\end{subtable}
\label{tab:3}
\begin{subtable}[t]{0.51\textwidth}
\centering
\begin{tabular}{@{}|l |c |c|@{}}     \hline
\rule{0pt}{1\normalbaselineskip} Method & SVHN 10k labels & CelebA 20k labels  \\    \hline \hline
\rule{0pt}{1\normalbaselineskip} CNN  & 16.45 $\pm$ 0.17 & 22.24 $\pm$ 0.12   \\                    
VAE & 16.81 $\pm$ 0.24 & 21.67 $\pm$ 0.35   \\  
STN \cite{jaderberg2015spatial} & 14.97 $\pm$ 0.16 & 18.82 $\pm$ 0.22   \\ 
GCNN \cite{singh2022learning} & 12.85 $\pm$ 0.26 & 16.98 $\pm$ 0.35   \\  
L-STN \cite{jiang2019linearized} & 11.45 $\pm$ 0.14 & 15.31 $\pm$ 0.21   \\  
SE3Mov \cite{sangalli2022moving} & 9.85 $\pm$ 0.27 & 13.67 $\pm$ 0.24   \\   
ScDCFNet \cite{zhu2022scaling} & 9.56 $\pm$ 0.09 & 13.63 $\pm$ 0.14   \\ \rowcolor{LightCyan}
 ESTN (ours) & {\bf 9.14 $\pm$ 0.24} & {\bf 13.24 $\pm$ 0.16}   \\   \hline
\end{tabular}
\end{subtable}
 \label{tab:4}
 }
 \end{table*}
\begin{table}[h]
\centering
\small{
\caption{The architecture of CNNs for experiments on Fashion MNIST, SVHN and CelebA. All Conv. and fully connected (FC) layers are followed by ReLU activations and batch normalization.}
\begin{tabular}{@{}|c |l |r|@{}}     \hline
\rule{0pt}{0.8\normalbaselineskip} F-MNIST network &  SVHN network  & CelebA network \\  [0.5ex]   \hline
\rowcolor{Gray}
\multicolumn{3}{|c|}{$32\times32$ image} \\   [0.5ex] \hline
\rule{0pt}{1\normalbaselineskip} $3\times3$ conv. 32 & $3\times3$ conv. 64 & $3\times3$ conv. 128 \\   [0.5ex]  \hline
\rule{0pt}{1\normalbaselineskip} $3\times3$ conv. 32 & $3\times3$ conv. 64 & $3\times3$ conv. 128 \\    [0.5ex]  \hline\hline
\rowcolor{Gray}
 \multicolumn{3}{|c|}{EST layer} \\  [0.5ex]  \hline 
$3\times3$ conv. 32& $3\times3$ conv. 64 & $3\times3$ conv. 128\\    [0.5ex]  \hline\hline
\rowcolor{Gray}
\multicolumn{3}{|c|}{$2\times2$ max-pool, stride 2 dropout} \\   [0.5ex]  \hline 
\rule{0pt}{1\normalbaselineskip} $3\times3$ conv. 64 & $3\times3$ conv. 128 & $3\times3$ conv. 256 \\   [0.5ex]  \hline
\rule{0pt}{1\normalbaselineskip} $3\times3$ conv. 64 & $3\times3$ conv. 128 & $3\times3$ conv. 256 \\  [0.5ex]    \hline\hline
\rowcolor{Gray}
 \multicolumn{3}{|c|}{EST layer} \\  [0.5ex] \hline
\rule{0pt}{1\normalbaselineskip} $3\times3$ conv. 64 & $3\times3$ conv. 128 & $3\times3$ conv. 256 \\ [0.5ex]     \hline\hline
 \rowcolor{Gray}
 \multicolumn{3}{|c|}{$2\times2$ max-pool, stride 2 dropout} \\ [0.5ex]  \hline 
\rule{0pt}{1\normalbaselineskip} $3\times3$ conv. 64 & $3\times3$ conv.128 & $3\times3$ conv. 512 \\  [0.5ex]  \hline
\rule{0pt}{1\normalbaselineskip} $1\times1$ conv. 64 & $1\times1$ conv. 128 & $1\times1$ conv. 256\\   [0.5ex] \hline\hline
\rowcolor{Gray}
 \multicolumn{3}{|c|}{EST layer} \\  [0.5ex]   \hline
\rule{0pt}{1\normalbaselineskip} $1\times1$ conv. 64  & $1\times1$ conv. 128 & $1\times1$ conv. 128 \\  [0.5ex]   \hline\hline
 \rowcolor{Gray}
 \multicolumn{3}{|c|}{global average pool, $6\times6\rightarrow1\times1$} \\  [0.5ex]   \hline
\rule{0pt}{1\normalbaselineskip} FC 64$\rightarrow$10 & FC 128$\rightarrow$ 10 & FC 128 $\rightarrow$ 10 \\   [0.5ex]   \hline
\end{tabular}
 \label{tab:2-1}
}
 \end{table}

\section{Experiments}
In this section, the performance of our ESTN model is evaluated on image reconstruction and classification, and compared with several state-of-the-art methods. Image reconstruction can be seen as a prediction of pixel intensities to determine a target pixel's location which is sensitive to the detected contextual points (pixels). In our experiments, the contextual set may be diverse but the target set consists of all the image pixels. For image reconstruction, the same model is utilized for 1-D function regression (excluding the setting of the last layer as 3-D information for RGB images). The models are evaluated on five public datasets: MNIST \cite{lecun1998gradient}, Fashion-MNIST \cite{xiao2017fashion}, SVHN \cite{netzer2011reading}, zero shot multi-MNIST (ZSMM) \cite{lu2017zero}, and CelebA \cite{liu2018large}. In all our experiments, we use a VAE architecture based on \cite{schott2018towards}. The encoder contains 4 layers (convolution sizes = $[32, 32, 64, 2\times8]$, strides = [1, 2, 2, 1], and kernel sizes = [5, 4, 3, 5]). ReLU activation function is used in the first three layers, whereas the final layer is linear. The decoder also contains 4 layers (convolution sizes = [32, 16, 16, 1], strides = [1, 2, 2, 1], and kernel sizes = [4, 5, 5, 3]). ReLU activation function is used in the first three layers, whereas the sigmoid activation function is used in the final layer, and batch normalization is used on all the layers except the last ones. To train our network, Adam is used to optimize the weights, with a learning rate of 0.0001, weight decay = 0.001, and momentum = 0.9. Our ESTN contains two ST layers that are trained together with the encoder and the decoder, and the loss is determined as the difference between the decoder’s output and the target. We implement the proposed ESTN in Pytorch, and run it on a Tesla P40 GPU. 

Predicting pixel values can be viewed as a regression problem that maps a 2D pixel position to its pixel intensity. In general, each image represents a single realization of the process that was sampled on a fixed two-dimensional grid. We use the regular train/test split with up to 250 contextual points (image pixels) to train the ESTN on the datasets. In Table \ref{tab:1}, we report the mean and standard deviations of log-likelihood of ESTN compare to different VAE-based models such as Conditional VAE (CVAE) and $\beta$-VAE.
CVAE \cite{ramchandran2022learning} is a modification of VAE to reconstruct an image conditioned on the given attributes and $\beta$-VAE \cite{chen2018isolating} is another variation of the VAE that is often used as an unsupervised approach for the representation learning.
Results indicate that ESTN performs considerably better than CVAE and $\beta$-VAE. Fig. \ref{fig:5} shows a visual representation of the marginal likelihoods of the images with a random number of contextual pixels, along with the distribution results of the ESTN, CVAE and $\beta$-VAE. Results show that ESTN has accurate contextual reconstruction performance compared to the other baselines. Moreover, Fig. \ref{fig:4} shows the performance of ESTN on qualitative samples with different noise set-ups. These results show that even when the target pixels are filled, ESTN generates smooth results due to the effective incorporation of transformer layers and manifold interpolation.
\begin{figure*}
\centering
\hspace{-6mm}
\includegraphics[width=1.92\columnwidth]{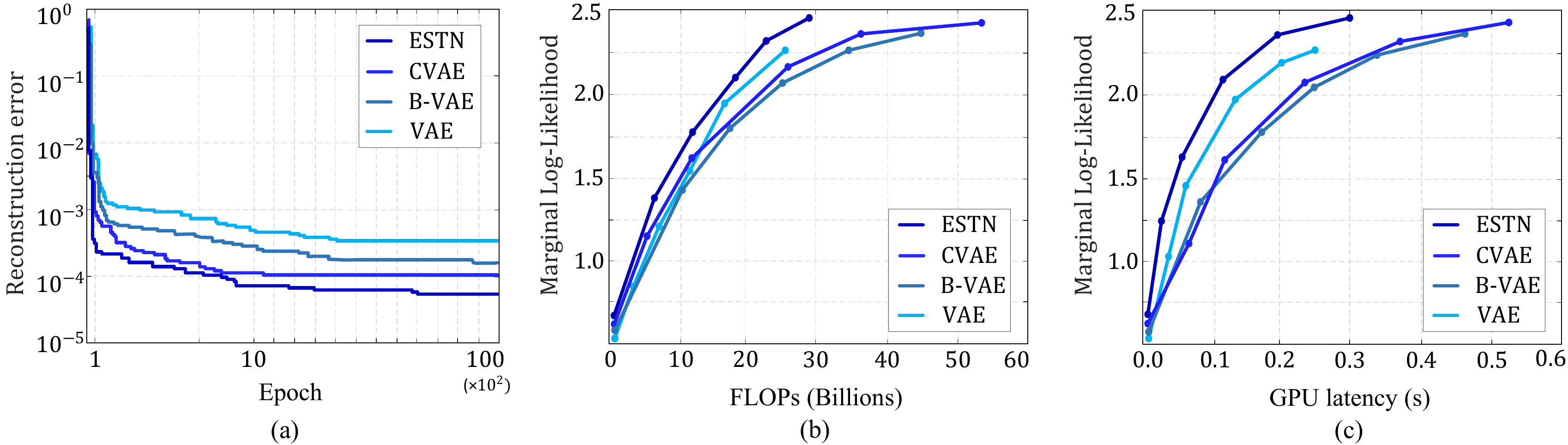} 
\caption{Model performance, efficiency and inference latency comparison on Fashion MNIST. Performance is measured with batch size 1 on the same machine. }   
\label{fig:8}
\end{figure*}
\subsection{Fashion MNIST}
For the classification task, we applied ESTN to learn the underlying data manifold.
To undertake experiments on the Fashion MNIST, we use three sets of the labeled data, {\it i.e.}, 1k, 3k and 5k, respectively and preserve 150 data points for validation from the initial training set. The network architecture for classification is reported in Table \ref{tab:2-1}. We set the batch size to 32. The model is trained for 5k epochs. Adam optimizer is adopted with the initial learning rate of 0.001, and linearly decayed on every 1k epochs. The training time was about 20 hours. The hyperparameters are tuned in the scale of the tangent perturbation where we use rotations, and scale changes with respect to the vertical-horizontal directions. We adopt Gaussian noise with standard deviation 1. As pre-processing, we normalized the pixels value into [-1, 1], standard deviation is set to $\pi/4$ for rotations, $\sqrt{2}$ for scale changes either in vertical or horizontal direction, and 0.2 for translations. 0.1 applied for all the other hyperparameters. In the VAEs network, for estimating the underlying manifold, six convolution layers and two fully connected layers are used, where the decoder is symmetric to the encoder, replacing the convolution layers with the deconvolution ones.

Table \ref{tab:3} reports the results. ESTN has lower classification errors as compared to STN and L-STN, in various numbers of the labeled data. The performance of ESTN is relying on the prediction of the underlying data manifold. Therefore, ESTN contributes better results compared to other learning-based equivariant CNNs.

\subsection{SVHN and CelebA}
We performed two sets of experiments to prove the success of ESTN in classification, SVHN with 10k labeled data, and CelebA with 20k labeled data. As pre-processing, we normalized the pixels value into [-1, 1]. The networks' structure is provided in Table \ref{tab:2-1}. We set the batch size to 32. The SVHN's network is trained for 8k epochs, and the CelebA's network is trained for 12k epochs. The initial learning rate is set to 0.001, and linearly decayed on every 2k epochs. The other hyperparameters are similar to the Fashion MNIST's setting. Table \ref{tab:3} shows the results of our experiments on SVHN and CelebA. The evaluation results indicate the superior performance of ESTN compared to the other state-of-the-art approaches. Particularly, in comparison with L-STN that used linear interpolation, ESTN shows significant improvements with very similar network settings. In addition, ESTN has higher accuracy compared to the recent developed methods such as ScDCFNet \cite{zhu2022scaling} and SE3Mov \cite{sangalli2022moving}.
\begin{table}[h]
\centering
\caption{Ablation experiments on three datasets.}
\small{
\begin{tabular}{@{}|l| c |c|c|@{}}  \hline  
\rule{0pt}{0.8\normalbaselineskip} Model & MNIST & F-MNIST & SVHN  \\    \hline\hline 
\rowcolor{LightCyan}
ESTN (ours) & 1.28 $\pm$ 0.01& 2.47 $\pm$ 0.00 & 3.95 $\pm$ 0.01    \\ [-0.1ex]  
 ESTN w/o norm. & 1.20 $\pm$ 0.03& 2.38 $\pm$ 0.02 & 3.84 $\pm$ 0.02    \\ [-0.1ex]                                 ESTN w/o ER & 1.22 $\pm$ 0.02& 2.41 $\pm$ 0.01 & 3.89 $\pm$ 0.00     \\ [-0.1ex]   
\hline
\end{tabular}
 \label{tab:5}
}
 \end{table}
\subsection{Ablation study}
We performed the ablation study to show the impact of each module on the overall performance of ESTN. The evaluation results are shown in Tables \ref{tab:5}. In our proposed model, both the entropy regularizer, and the norm-regularization methods are extremely helpful to improve the performance, otherwise the network fails to impose the manifold invariance or cannot be robust against the off-manifold noise. 
Our experiments illustrate that sampling based on the underlying data manifold can significantly improves the performance of ESTN. ESTN performs satisfactorily with a wide range of manifold coordinate charts. In our experiments, CNN produces more detailed images than VAEs, on the other hand, VAEs actively cooperates with ESTN to produce more realistic samples. It has been observed that sampling based on tangent bundle data manifold helps ESTN to explore high-dimensional data by characterizing the learning task. 

Fig. \ref{fig:8} (a) shows, ESTN has much more rapid decreases in reconstruction error and lower values at convergence. Compare to the other baselines ESTN has around $10^{-2}$ lower reconstruction errors. In Fig. \ref{fig:8} (b), we evaluate the efficiency of ESTN in comparison with that of the other models. The results indicate that ESTN can considerably enhance the system efficiency. For instance, ESTN surpasses CVAE and $\beta$-VAE . This suggests that we can have better performance by interpolating based on the underlying data manifold, and capture the attention to the significance of norm-regularization by modifying a small part of the network.
Fig. \ref{fig:8} (c) demonstrate the GPU latencey-accuracy curves for the models, where ESTN achieves higher accuracy in shorter time with much fewer FLOPS against the other models. Our ESTN is computational cheaper, for example, up to 2.1$\times$ faster than CVAE.

\section{Conclusion} 
In this paper, we have proposed a new variant of spatial transformer networks based on sampling on the tangent space to explore more samples in the search space along the data manifold to improve the gradients and preserve the norm for differentiable image handling. We have demonstrated that ESTN have been successful on a wide range of challenges and is highly scalable. In terms of the arithmetic operations in the vector space, our tangent space manifold convolution is sensible. In fact, the proposed ESTN can be used as an example to improve the gradient estimation with better performance on image reconstruction and classification tasks. Moreover, since the proposed model preserves the norm of gradients, it can be used in highly deep networks and reduces the optimization bottlenecks of deep networks.

{\small
\bibliographystyle{ieee_fullname}
\bibliography{egbib}
}

\end{document}